\pdfoutput=1

\documentclass[11pt]{article}

\usepackage[preprint]{acl}

\usepackage{natbib}
\usepackage{times}
\usepackage{latexsym}
\usepackage{microtype}
\usepackage{makecell}
\usepackage{subfigure}
\usepackage{booktabs} 
\usepackage{algorithm}
\usepackage{algorithmic}
\usepackage{amsmath}
\usepackage{amsfonts}
\usepackage{multirow}
\usepackage{array}
\usepackage{longtable}

\usepackage[T1]{fontenc}

\usepackage[utf8]{inputenc}

\usepackage{microtype}

\usepackage{inconsolata}

\usepackage{graphicx}
\usepackage{hyperref}
\title{Align Attention Heads Before Merging Them: An Effective Way for Converting MHA to GQA}

\author{
 \textbf{Qingyun Jin\textsuperscript{1,2}}
 \qquad
 \textbf{Xiaohui Song\textsuperscript{2}}
 \qquad
 \textbf{Feng Zhou\textsuperscript{2}}
 \qquad
 \textbf{Zengchang Qin\textsuperscript{1} \thanks{Corresponding author.}}
\\
 \textsuperscript{1}Beihang University, Beijing, China
 \\
 \textsuperscript{2}OPPO AI Center, Beijing, China
\\
 \texttt{\{jinqingyun, zcqin\}@buaa.edu.cn}
\\
 \texttt{\{songxiaohui, zhoufeng1\}@oppo.com}
}

\begin{document}
\maketitle
\begin{abstract}
Large language models (LLMs) have demonstrated exceptional performance across diverse natural language processing tasks. However, as the model size and the input sequence's length increase, the linearly increasing key-value (KV) cache significantly degrades inference throughput. Therefore, grouped-query attention (GQA), as an alternative to multi-head attention (MHA), has been widely introduced into LLMs. In this work, we propose a cost-effective method for converting MHA into GQA with any compression ratio of KV heads. The key point of our method lies in the application of Procrustes analysis to the attention heads, which enhances the similarity among attention heads while preserving computational invariance, thereby improving the model's post-training performance. Subsequently, we employ $\mathit{L_0}$ regularization to prune redundant parameters. The model after pruning can be adapted to the standard GQA framework. Experimental results show that our strategy can compress up to 87.5\% KV heads of LLaMA2-7B model and 75\% KV heads of Sheared-LLaMA-1.3B with acceptable performance degradation. Our code is released at \href{https://github.com/fpcsong/mha2gqa}{https://github.com/fpcsong/mha2gqa}.
\end{abstract}

\section{Introduction}
Recently, large language models (LLMs) \cite{radford2018improving, NEURIPS2020_1457c0d6, ouyang2022training} show remarkable performance on a variety of natural language processing tasks. However, since most LLMs are based on Transformer architecture \cite{NIPS2017_3f5ee243}, the expansion of the sequence length during inference leads to a linear increase in the memory footprint of key-value (KV) cache, which substantially increases on-device memory consumption. A reduced KV cache footprint not only lowers inference costs but also facilitates the processing of longer sequences and improves inference speed. Therefore,  reducing the size of the KV cache is a key issue for LLMs.

Many methods for KV cache compression have been proposed, including KV cache quantization \cite{hooper2024kvquant, yue2024wkvquant, yang2024no}, tokens dropping \cite{adnan2024keyformer, liu2023scissorhands, tang2024razorattention} and so on. However, these approaches often introduce additional computational procedures, which are incompatible with general LLM frameworks.

Another approach to KV cache compression is to directly change the attention architecture. Multi-query attention (MQA) \cite{shazeer2019fast} and grouped-query attention (GQA) \cite{ainslie2023gqa} reduce KV cache by allowing multiple attention heads to share a single KV head, offering a simple and effective solution for attention optimization. Since GQA has better inference stability and performance, it has been widely used in LLaMA 2 \cite{touvron2023llama}, LLaMA 3 \cite{dubey2024llama}, Qwen2 \cite{yang2024qwen2}, Mistral \cite{jiang2023mistral} and other LLMs \cite{liu2024sora, zhang2024tinyllama}. Multi-head latent attention (MLA) \cite{liu2024deepseek-v2} further reduces KV cache through low-rank projection of the cached data. MLA has been used in the DeepSeek model series \cite{liu2024deepseek-v2,liu2024deepseek-v3}. These efficient attention architectures achieve KV cache compression with better universality. 

In this study, we present a method that converts MHA to GQA to compress KV cache. Inspired by the idea of computational invariance in LLMs \cite{ashkboos2024slicegpt} and Procrustes analysis \cite{schonemann1966generalized}, we apply proper orthogonal transformations to the projection matrices in attention heads to simplify the conversion from MHA to GQA: Specifically, we regroup attention heads based on the similarity of their KV caches and use generalized Procrustes analysis \cite{enwiki:1126373270} to maximize the similarity of attention heads within each group. This transformation preserves the model's output invariance. Finally, $\mathit{L_0}$ regularization \cite{louizos2017learning} is applied to transfer original KV heads to new ones to get GQA. Figure~\ref{fig:process} illustrates the overall framework of our method. Experimental results show that attention head transformation can significantly improve the performance of the pruned model. Our contributions are as follows.

\begin{itemize}
\item Based on the idea of computational invariance, we employ Procrustes analysis to enhance the similarity among attention heads. This approach not only improves the performance of GQA model, but also provides a new perspective for evaluating the similarity between attention heads, offering new insights for future research related to compressing KV cache.

\item We propose a general and cost-effective method for converting MHA to GQA, using $\mathit{L_0}$ regularization to compress the key and value heads to any percentage and basically restore performance after supervised fine-tuning. 

\item We conduct experiments on LLaMA2-7B \cite{touvron2023llama} and Sheared-LLaMA-1.3B \cite{xia2023sheared}, converting them into GQA models of varying sizes, separately. Model performance does not decrease significantly compared to that of MHA model.
\begin{figure*}[t]
  \includegraphics[trim=1cm 4.8cm 2cm 1cm, clip, width=\linewidth]{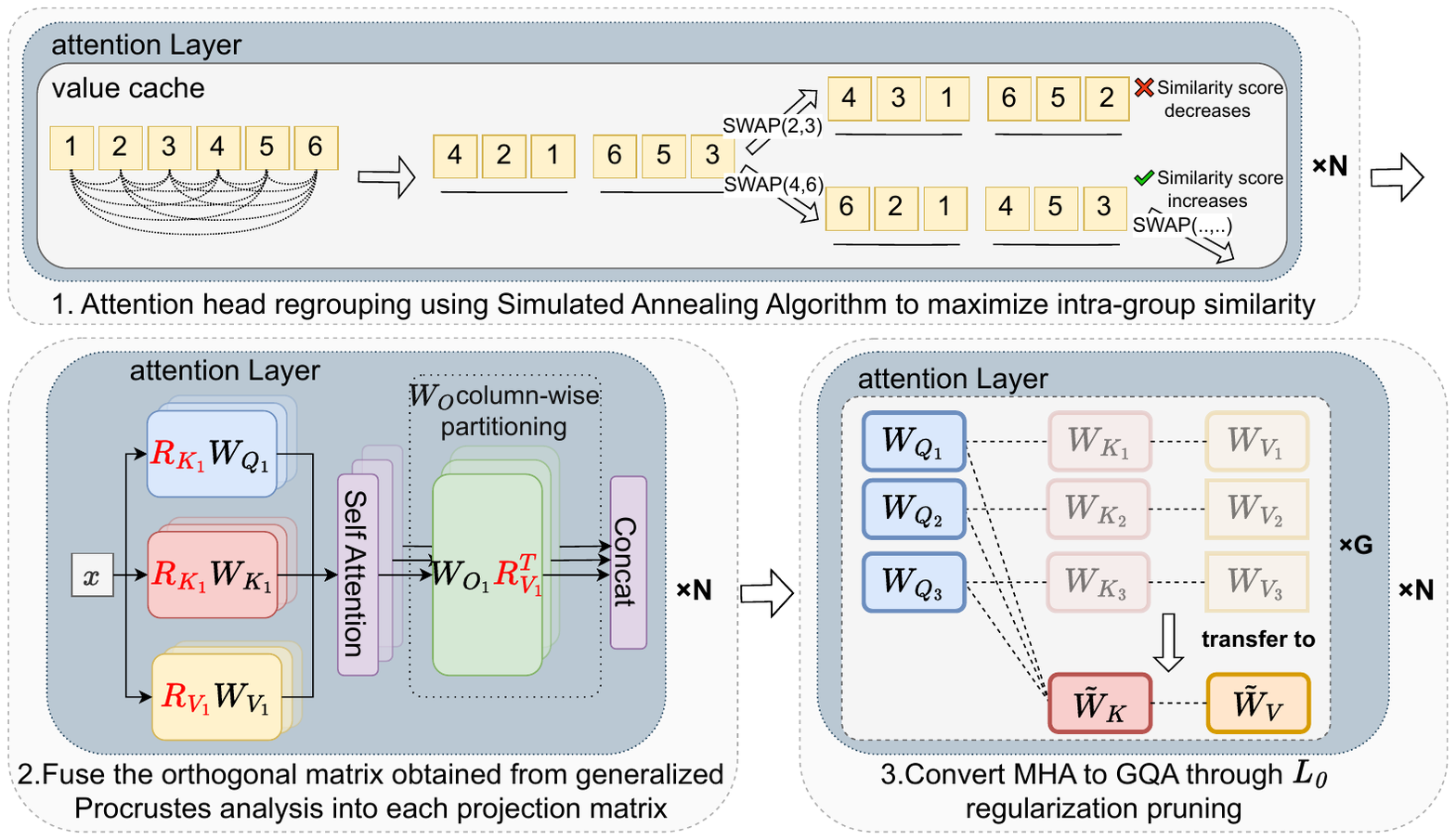}
  \caption {An illustration of our method. 1. In each attention layer of LLM, after the average similarity score is evaluated by Procrustes analysis between every two value caches (or key caches). Use Simulated Annealing Algorithm to search the optimal grouping result that maximizes the similarity score among value caches (or key caches) in each layer. 2. After grouping, then fuse the orthogonal matrix obtained from generalized Procrustes analysis into each projection matrix to enhance the similarity among attention heads in each group without changing the model. 3. During training, \(\mathit{L_0}\) loss is used to gradually transfer original key-value projection matrices to newly added ones within each group (there are \(G\) groups in one attention layer).  After pruning, original key-value projection matrices will be discarded, then we get a standard GQA model.
}
  \label{fig:process}
\end{figure*}
\end{itemize}

\section{Related Works}
\subsection{\texorpdfstring{\(\mathit{L_0}\)}{L0} regularization}
\(\mathit{L_0}\) regularization \cite{louizos2017learning} is a structured pruning approach that transforms a pruning problem into an optimization problem under constraints. The pruning process is performed simultaneously with model optimization by introducing trainable masks. With the wide application of LLMs, this method has been applied to compressing LLMs. In the work of \cite{wang2019structured}, the $\mathit{L_0}$ method is applied based on low-rank pruning to achieve further improvements in effectiveness, and they propose to gradually increase the target size at a linear rate during the process of pruning training. In CoFi \cite{xia2022structured}, the $\mathit{L_0}$ method is applied directly to LLMs by introducing pruning masks with different granularities. They prune the hidden dimension, the intermediate dimension, the number of attention heads, and even an entire MHA or FFN layer. The subsequent work Sheared-LLaMA \cite{xia2023sheared} incorporates previous methods and specifies the target structure so that the pruned model can be directly adapted to standard LLM frameworks.

\subsection{Transfer MHA to GQA}
\cite{ainslie2023gqa} proposes GQA for the first time, in which MHA is converted to GQA using mean pooling initialization. However, this method requires uptraining to restore performance and incurs significant computational costs. \cite{yu2024effectively} keeps the corresponding parameters based on the principal components of the collected KV caches, then uses LoRA \cite{hu2021lora} to fine-tune the model to restore performance. \cite{chen2024optimised} proposes to regroup attention heads based on the criterion of cosine similarity and allows for varying group sizes. DHA\cite{chen2024dha} adaptively configures group sharing for key heads and value heads across various layers, achieving a better balance between performance and efficiency. However, none of the aforementioned improvement methods can be fully adapted to the standard GQA model.

\subsection{Compressing model based on the principal components of features}
Some previous works \cite{liu2023deja,yu2023compressing} have pointed out that the features of LLMs are generally low-rank. Therefore, identifying and deleting the low-rank components of the model is an effective method for model compression. 

Low Rank BERT \cite{noach2020compressing} reduces the number of parameters and increases inference speed by decomposing the weight matrices into two low-rank matrices. SliceGPT \cite{ashkboos2024slicegpt} introduces the idea of computational invariance in Transformer architecture and removes columns or rows of the transformed weight matrices to reduce model size. \cite{yu2024effectively} applies orthogonal transformations to key-value projection matrices by analyzing the low-rank features of KV cache.

\section{Method}
In this section, we will specifically describe our method. Our method consists of two parts, transformation of attention heads and pruning training. Transformation of attention heads represents employing Procrustes analysis to align projection matrices in order to enhance the similarity between attention heads of the same group, so that we can increase efficiency of merging attention heads. The pruning training process combines pruning with $\mathit{L_0}$ regularization \cite{louizos2017learning} and knowledge distillation \cite{gou2021knowledge}.

\subsection{Motivation}
\begin{figure*}[t]
  \includegraphics[trim=3cm 2.3cm 2cm 1.5cm, clip, width=\linewidth]{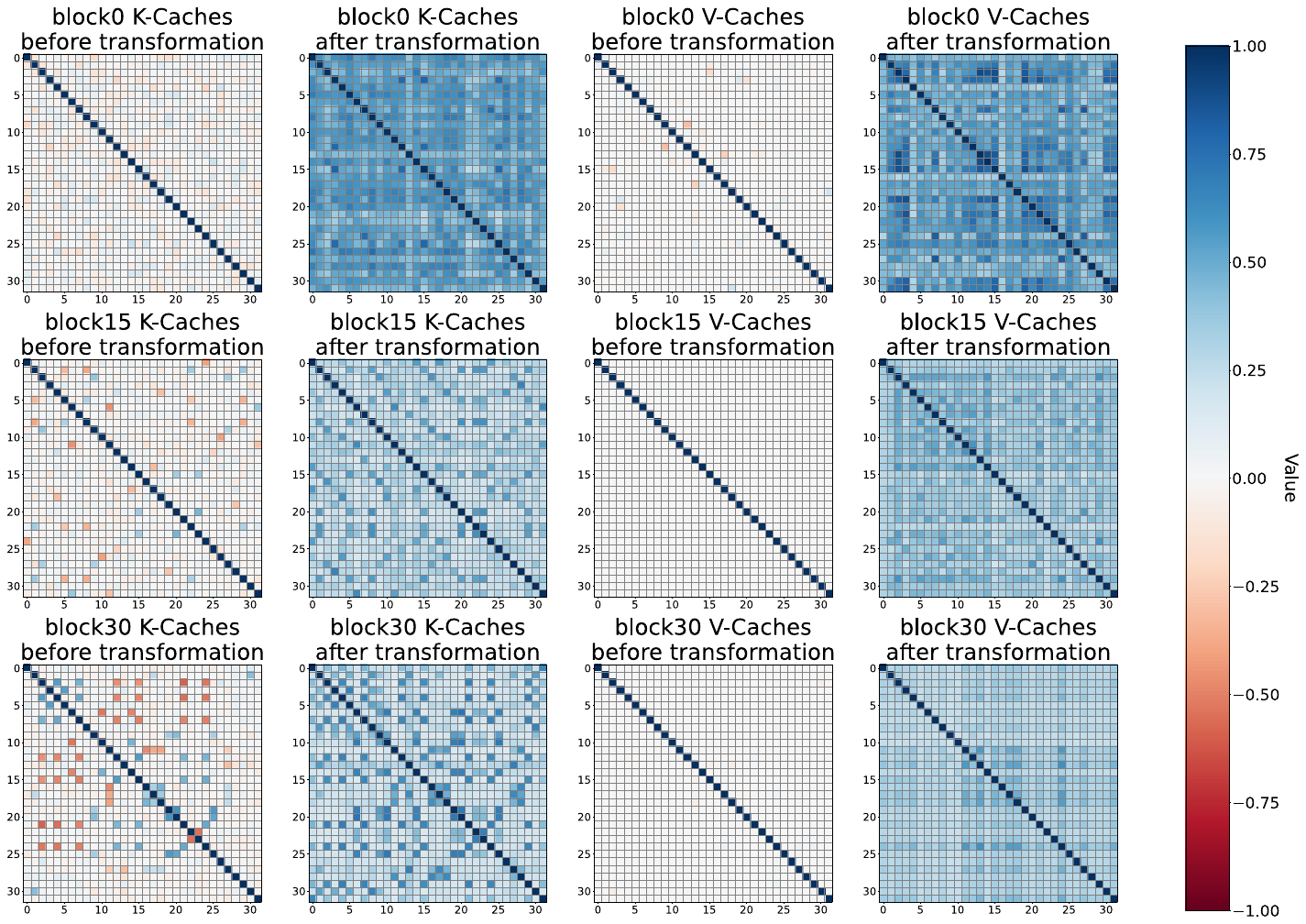}
  \caption {This figure shows the average cosine similarity of key and value caches between any two heads before and after applying transformation in some blocks of LLaMA2-7B. It can significantly improve the cosine similarity between KV caches after Procrustes analysis (In fact, it can also significantly reduce the Euclidean distance between any two caches.).}
  \label{fig:B-fig3}
\end{figure*}

To analyze the characteristics of KV cache, we follow a prior calibration method for LLMs \cite{frantar2023sparsegpt,sun2023simple} in order to obtain calibration data: Sample 128 sequences from the C4 \cite{raffel2020exploring} training set and each sequence is 2048 tokens long, 262144 tokens in total. Then perform model inference on LLaMA2-7B and collect KV caches, i.e.,
\begin{equation}
K=[K_1;\ldots;K_H] \quad \quad V=[V_1;\ldots;V_H]
\end{equation}
where \(K,V\in\mathbb{R}^{d\times N}\) are KV caches corresponding to each block, which can be divided into \(K_i,V_i\in\mathbb{R}^{d_H\times N}\), \(N\) is the number of tokens, \({d}\) is embedding dimension and \({H}\) represents the number of heads in each MHA, \({d_H}\) is set to \({d/H}\), then we can calculate the average cosine similarity between each of two heads as follows:
\begin{equation}
SimK_{i,j}^{ori}=\frac{1}{N}\sum_{n=0}^{N-1}cos({{K}_i[n]\cdot {K}_j[n]})
\end{equation}
\begin{equation}
SimV_{i,j}^{ori}=\frac{1}{N}\sum_{n=0}^{N-1}cos({{V}_i[n]\cdot {V}_j[n]})
\end{equation}
where \(i,j\) are any two of attention heads in the same block, \(n\) represents the \(n^{th}\) token in this cache. Taking LLaMA2-7B as an example, as shown in Figure~\ref{fig:B-fig3}, we notice that the vast majority of KV caches are almost orthogonal. 

However, according to \cite{yu2024effectively}, KV caches are low-rank. Given that these caches occupy only a portion of spatial dimensions, applying appropriate orthogonal transformations to the projection matrices to align key and value caches can increase efficiency of merging attention heads. Fortunately, this approach is feasible.

\subsection{Preliminaries} 
Given two sets of vectors of the same shape: \( \mathbf{X} = \{ \mathbf{x}_1, \mathbf{x}_2, \ldots, \mathbf{x}_N \} \in \mathbb{R}^{M \times N} \) and $\mathbf{Y} = \{ \mathbf{y}_1, \mathbf{y}_2, \ldots, \mathbf{y}_N \} \in \mathbb{R}^{M \times N}$, how to find the optimal orthogonal transformation that aligns the two sets of vectors? This kind of problems is called Orthogonal Procrustes problem, and its mathematical formulation is as follows:
\begin{equation}
\min_{\mathbf{Q}} \|\mathbf{Q}\mathbf{X}- \mathbf{Y}\|_F^2
\end{equation}
The optimal orthogonal transformation can be derived from SVD of the matrix \( \mathbf{Y} \mathbf{X}^T\), the general solution is
\cite{schonemann1966generalized}:

Perform SVD on the covariance matrix of \textbf{X} and \textbf{Y},
\begin{equation}
(\mathbf{Y} \mathbf{X}^T) = \mathbf{\Phi} \mathbf{\Sigma} \mathbf{\Psi}^T
\end{equation}
Then obtain the optimal orthogonal matrix $\mathbf{Q}$:
\begin{equation}
\mathbf{Q} = \mathbf{\Psi} \mathbf{\Phi}^T
\end{equation}

We can use the same way to align any two KV caches from different attention heads in the same block. Furthermore, if we want to align more than two sets of caches, generalized Procrustes analysis \cite{enwiki:1126373270} is a good solution. The detailed description is shown in algorithm~\ref{alg:GPA}. 
\begin{algorithm}[h]
\caption{Generalized Procrustes Analysis}
\label{alg:GPA}
\begin{algorithmic}
\REQUIRE Matrices $X_1, X_2, \ldots, X_H$
\ENSURE Aligned matrices $Y_1, Y_2, \ldots, Y_H$

\STATE Initialize $Y_i = X_i$ for all $i$
\STATE Compute mean shape $\bar{M} = \frac{1}{H} \sum_{i=1}^{H} Y_i$

\REPEAT
    \FOR{$i = 1$ to $H$}
        \STATE Compute $\Phi_i \Sigma_i \Psi_i^T = \text{SVD}(Y_i^T M)$
        \STATE Update $Y_i = Y_i \Psi_i \Phi_i^T$
    \ENDFOR
    \STATE Update mean shape $\bar{M} = \frac{1}{H} \sum_{i=1}^{H} Y_i$
\UNTIL{convergence}
\RETURN $Y_1, Y_2, \ldots, Y_H$
\end{algorithmic}
\end{algorithm}
\subsection{Transformation of attention heads}
\label{subsec:standard}
To calculate the optimal orthogonal matrix for each pair of key and value heads, we collect KV caches according to the method mentioned above. Here, we use two criteria to perform Procrustes analysis. 

\textbf{Based on cosine similarity.} Firstly normalize each vector in \(K_i\) and \(V_i\) to roughly reduce the influence of magnitude of the vector:
\begin{equation}
\hat{K}_i[*] = \frac{K_i[*]}{\|K_i[*]\|}
\end{equation}
\begin{equation}
\hat{V}_i[*] = \frac{V_i[*]}{\|V_i[*]\|}
\end{equation}
then we can get the optimal orthogonal matrix \(Q_V\) to align any two value caches, taking \(\hat{V_i}\) and \(\hat{V_j}\) as example:
\begin{equation}
   (\hat{V_i} \hat{V_j}^T) = \Phi \Sigma \Psi^T
\end{equation}
\begin{equation}
   Q_{V_j}=\Psi\Phi^T
\end{equation}
For each block, the output of self-attention layer can be seen as the sum of all attention heads:

\begin{small}
\begin{equation}
\begin{aligned}
   \text{MultiHead}(W_Q, W_K, W_V, W_O)\\ = \sum_{i=1}^{H} (W_{O_i}(W_{V_i}X)\text{Softmax}\left(\frac{(W_{K_i}X)^T (W_{Q_i}x)}{\sqrt{d_H}}\right))
\end{aligned}
\end{equation}
\end{small}
where the projection matrices in the attention heads are \(W_{Q_i}, W_{K_i}, W_{V_i}\in\mathbb{R}^{d_{H}\times d}\) and \(W_{O_i}\in\mathbb{R}^{d\times d_{H}}\), \(X\in\mathbb{R}^{d\times len}\) represents the previous tokens, and \(x\in\mathbb{R}^{d\times 1}\) represents the current token. For brevity, Rotary position embedding (RoPE) is ignored here. Then we can fuse the orthogonal matrix into the value projection matrix ${W}_{V_j}$ and the output projection matrix $W_{O_j}$ to ensure computational invariance:
\begin{equation}
   W^{'}_{V_j} = Q_{V_j} W_{V_j}
\end{equation}
\begin{equation}
   W^{'}_{O_j} =W_{O_j} Q_{V_j}^T 
\end{equation}
As for \(W_Q\) and \(W_K\), due to the existence of RoPE \cite{su2024roformer}, Procrustes analysis cannot be applied directly. However, we can divide the d-dimension space into \(d/2\) sub-spaces and apply Procrustes analysis in every two dimension just like RoPE, which is to say the orthogonal matrix for key projection matrix should be in this form:
\begin{equation}
R_{K_j} = \begin{pmatrix}R_{\theta_1} & 0 & \cdots & 0 \\0 & R_{\theta_2} & \cdots & 0 \\\vdots & \vdots & \ddots & \vdots \\0 & 0 & \cdots & R_{\theta_{d/2}}
\end{pmatrix}
\end{equation}
where \(R_{\theta_{\cdot}}\) is a 2D rotation matrix. Then we fuse the orthogonal matrix \(R_{K_j}\) into the query projection matrix $W_{Q_j}$ and key projection matrix $W_{K_j}$:
\begin{equation}
   W^{'}_{Q_j} = R_{k_j} W_{Q_j}
\end{equation}
\begin{equation}
   W^{'}_{K_j} = R_{k_j} W_{K_j} 
\end{equation}
So, we get
\begin{small} 
\begin{equation}
\begin{aligned}
q_s^{'T} k^{'}_{t}&=(R_{\Theta,s}^d (R_{K_j} {W}_{Q_j}) {x}_s)^{T}({R}_{\Theta,t}^d (R_{K_j} {W}_{K_j}) {x}_t) \\ 
&={x}_s^{T}{W}_{Q_j}^{T} (R_{K_j}^T {R}_{\Theta,t-s}^d R_{K_j}) {W}_{K_j} {x}_t \\
&= {x}_s^{T} {W}_{Q_j}^{T} {R}_{\Theta,t-s}^d {W}_{K_j} {x}_t \\
&=(R_{\Theta,s}^d {W}_{Q_j} {x}_s)^{T}({R}_{\Theta,t}^d  {W}_{K_j} {x}_t) \\ 
&={q}_s^T {k}_{t}
\end{aligned}
\end{equation}
\end{small} 
where \(q_s\) represents the query of the \(s^{th}\) position and \(k_t\) represents the key of the \(t^{th}\) position. This transformation doesn't change the model either.

In this way, given any two key or value caches, we can use this method to calculate the maximum cosine similarity achievable.
\begin{equation}
SimK_{i,j}^{after}=\frac{1}{N} \sum_{n=0}^{N-1}cos({{K}_i[n]\cdot (R_{K_j}{K}_j[n])})
\end{equation}
\begin{equation}
SimV_{i,j}^{after}=\frac{1}{N} \sum_{n=0}^{N-1}cos({{V}_i[n]\cdot (Q_{V_j}{V}_j[n])})
\end{equation}
Noticing $SimV_{i,j}^{after}$ is equal to $SimV_{j,i}^{after}$, so is $SimK^{after}$. Figure~\ref{fig:B-fig3} shows the cosine similarity between KV caches before and after applying the transformation.

\textbf{Based on Euclidean distance.} Similar to applying transformations based on cosine similarity, we also apply transformations based on Euclidean distance between two KV caches. In this case, we don't normalize vectors and the similarity between two caches can be described as the negative value of the Euclidean distance of them, for brevity, only some key formulas are displayed here:
\begin{equation}
   ({V_i} {V_j}^T) = \Theta \Lambda \Omega^T
\end{equation}
\begin{equation}
   P_{V_j}=\Omega\Theta^T
\end{equation}
\begin{equation}
SimV_{i,j}^{after}=-\frac{1}{N} \sum_{n=0}^{N-1}\|{V}_i[n]- (P_{V_j}{V}_j[n])\|_F
\end{equation}
In the next section, we will compare the performance of the two criteria.
\subsection{Find better grouping method}
\label{subsec:grouping}
After obtaining the similarity scores between every pair of attention heads, we can regroup attention heads based on these scores. We define the similarity score of a group as the sum of similarity scores between every pair of attention heads within that group, and the total similarity score for each grouping result is the sum of similarity scores of all groups. Our objective is to identify the grouping result with the highest total score\footnote{While the highest similarity between pairs within the same group does not necessarily equate to the lowest cost in terms of converging to the same parameters during pruning, this strategy remains acceptable when considering computation and time costs.}. We use \(SimK^{after}\) and \(SimV^{after}\) as grouping criteria, respectively. In the next section, we will compare the performance of the two criteria. The mathematical expression of the score of a grouping result \(A=\{A_1, A_2,\cdots, A_G\}\) is as follows:
\begin{equation}
Score_{key}(A) = \sum_{g=1}^{G} \sum_{0 \leq i < j < D} SimK_{A_g[i],A_g[j]}^{after}
\end{equation}
\begin{equation}
Score_{value}(A) = \sum_{g=1}^{G} \sum_{0 \leq i < j < D} SimV_{A_g[i],A_g[j]}^{after}
\end{equation}
where \( A_g \) is the \( g^{th} \) group in \( G \) groups, the elements in \( A_g \) are the serial number of an attention head and there are \(D=H/G\) heads in a group.

We use Simulated Annealing Algorithm to get the best grouping result: Swap two random heads in different groups and calculate the new score, accepting the new result if it reaches a higher score. Repeat this process for multiple iterations. Because initialization has a significant impact on the final result, we run the algorithm multiple times. The details of the algorithm~\ref{alg:anneling} are shown below.
\begin{algorithm}[ht]
\caption{Simulated Annealing}
\label{alg:anneling}
\begin{algorithmic}
\REQUIRE $maxIter$, $epoch$, $SimV$ (or $SimK$)
\ENSURE grouping result with the highest score $bestG_n$
\STATE Set $score_{best}$ $\gets - \infty$
\FOR{$i = 1$ \text{ to } $epoch$}
    \STATE Initialize solution $G_n$ randomly
    \STATE $score_{current}$ $\gets$ calculate\_score($G_n$, $SimV$)
    \IF{$score_{current}$ $>$ $score_{best}$}
        \STATE Set $score_{best}$ $\gets$ $score_{current}$
        \STATE Set $bestG_n$ $\gets$ $G_n$
    \ENDIF
    \FOR{$j = 1$ \text{ to } $maxIter$}
        \STATE $G^{'}_n$ $\gets$ Randomly swap two elements from different groups in $G_n$
        \STATE $score_{new}$ $\gets$ calculate\_score($G^{'}_n$, $SimV$)
        \IF{$score_{new}$ $>$ $score_{current}$}
            \STATE Set $G_n$ $\gets$ $G^{'}_n$
            \STATE $score_{current}$ $\gets$ $score_{new}$
            \IF{$score_{new}$ $>$ $score_{best}$}
                \STATE Set $score_{best}$ $\gets$ $score_{new}$
                \STATE Set $bestG_n$ $\gets$ $G_n$
            \ENDIF
        \ENDIF
    \ENDFOR
\ENDFOR
\RETURN $bestG_n$
\end{algorithmic}
\end{algorithm}

After grouping, we can use Generalized Procrustes analysis to align attention heads in the same group.

\begin{table*}[ht]
  \centering
  \resizebox{\textwidth}{!}{
  \renewcommand\arraystretch{1.2}
  \begin{tabular}{c|llccccccccc|c}
    \hline
    \textbf{Model}&\multicolumn{2}{c}{\textbf{Methods}} & BoolQ & PIQA & HellaSwag & WinoGrande & ARC-C & ARC-E & OpenbookQA & SIQA & \textbf{Avg.} & \makecell[c]{\textbf{Budgets}\\(tokens and epochs)}\\
    \hline
    MHA & \multicolumn{2}{c}{Teacher} & 89.42& 77.15& 87.62& 85.16& 73.91& 85.09& 82.00& 78.10& 85.48& \makecell[c]{37.2M\\2 epochs}\\
    \hline
    \hline
    \multirow{7}{*}{GQA-16}&\multicolumn{2}{c}{baseline} & 86.85& 81.39& 87.08& 82.95& 66.22& 80.88& 79.00& 77.53& 84.60& \multirow{7}{*}{\makecell[c]{93M\\5 epochs}}\\
    \cline{2-12}
    &\multirow{2}{*}{default grouping} &cos & \textbf{88.62}& 81.50& 89.08& \textbf{83.82}& 67.56& 81.40& \textbf{81.40}& 77.53& 86.08&\\
    & &dist & 87.98& 81.61& 89.42& 83.14& 68.23& \textbf{82.63}& 79.20& \textbf{78.04}& 86.16&\\
    \cline{2-12}
    &\multirow{2}{*}{grouping by key} &cos & 88.47& 80.79& 88.26& 82.64& 69.23& 82.28& 79.20& 77.94& 85.53&\\
    & &dist & 88.44& 80.25& 88.96& 82.72& 70.23& 80.35& 77.40& 77.28& 85.68&\\
    \cline{2-12}
    &\multirow{2}{*}{grouping by value} &cos & 87.61& 81.23& 88.39& 82.79& 69.23& 80.35& 75.80& 77.12& 85.28&\\
    & &dist & 87.80& \textbf{82.10}& \textbf{89.66}& 83.50& \textbf{71.24}& 81.93& 80.80& 77.74& \textbf{86.35}&\\
    \hline
    \hline
    \multirow{7}{*}{GQA-8}&\multicolumn{2}{c}{baseline} & 84.56& 79.71& 83.53& 80.90& 60.54& 75.26& 75.00& 76.25& 81.65& \multirow{7}{*}{\makecell[c]{186M\\10 epochs}}\\
    \cline{2-12}
    &\multirow{2}{*}{default grouping} &cos & 86.76& 80.52& 86.66& 81.06& 64.88& 79.30& 77.60& \textbf{76.92}& 84.01&\\
    & &dist &86.76&\textbf{81.61}&\textbf{87.25}&82.64&65.22&\textbf{81.05}&77.80&76.61&\textbf{84.54}&\\
    \cline{2-12}
    &\multirow{2}{*}{grouping by key} &cos & \textbf{86.91}& 80.68& 87.01& 82.56& 64.21& 80.00& 76.60& 76.20& 84.24&\\
    & &dist &86.79&80.03&86.39&82.56&65.89&80.00&78.40&76.46&83.94&\\
    \cline{2-12}
    &\multirow{2}{*}{grouping by value} &cos & 86.39& 76.28& 81.80& 79.64&63.54&74.74&69.20&73.90&80.32&\\
    & &dist &86.60&81.50&86.96&\textbf{83.74}&\textbf{67.22}&79.47&\textbf{79.00}&76.31&84.42&\\
    \hline
    \hline
    \multirow{7}{*}{GQA-4}&\multicolumn{2}{c}{baseline} &81.86&76.93&76.97&78.30&55.52&73.86&69.80&74.56&77.03& \multirow{7}{*}{\makecell[c]{279M\\15 epochs}}\\
    \cline{2-12}
    &\multirow{2}{*}{default grouping} &cos &85.47&78.89&83.18&81.53&59.53&77.02&74.40&75.49&81.53&\\
    & &dist & 84.83& 79.27& 83.72& 80.74&59.53&77.54&\textbf{76.80}&75.54&81.77&\\
    \cline{2-12}
    &\multirow{2}{*}{grouping by key} &cos &85.41&79.38&83.37&80.90&61.20&77.19&74.40&75.49&81.66&\\
    & &dist &85.26&78.73&81.14&81.21&62.54&75.79&73.20&75.18&80.37&\\
    \cline{2-12}
    &\multirow{2}{*}{grouping by value} &cos &\textbf{86.18}&79.38&84.05&\textbf{82.16}&60.87&76.67&74.00&75.44&82.17&\\
    & &dist &85.69&\textbf{79.54}&\textbf{84.32}&82.00&\textbf{63.21}&\textbf{77.89}&75.40&\textbf{75.64}&\textbf{82.36}&\\
    \hline
    \hline
  \end{tabular}
  } 
   \caption{
    Performances of LLaMA2-7B with various methods.
    \label{tab:results1}
  }
\end{table*}
\subsection{Adaptation of \texorpdfstring{\(\mathit{L_0}\)}{L0} regularization}
During pruning training, we add new projection matrices which are initialized by mean-pooling all the original heads within that group to the model \cite{ainslie2023gqa}, here we use \(\tilde {W}_{K_{b, g}}\) or \(\tilde {W}_{V_{b, g}}\) to represent new projection matrices of the \( g^{th} \) group in the \( b^{th} \) block of the model:
\begin{equation}
\tilde {W}_{K_{b, g}} = \frac{1}{D}\sum^{D}_{i=1}W_{K_{b, (g-1)*D+i}}
\end{equation}
\begin{equation}
\tilde {W}_{V_{b, g}} = \frac{1}{D}\sum^{D}_{i=1}W_{V_{b, (g-1)*D+i}}
\end{equation}
These new matrices will be trained together with the model and replace original KV heads after pruning. Assume the model has \(N_{blocks}\) blocks and \(H\) heads in an attention layer, we introduce $\mathit{L_0}$ masks $z\in\mathbb{R}^{N_{blocks}\times H}$ \cite{louizos2017learning} to achieve this goal:
\begin{equation}
W_{K_{b, j}}^{apply} = z_{b,j}W_{K_{b, j}}+(1-z_{b,j})\tilde {W}_{K_{b, g}}
\end{equation}
\begin{equation}
W_{V_{b, j}}^{apply} = z_{b,j}W_{V_{b, j}}+(1-z_{b,j})\tilde {W}_{V_{b, g}}
\end{equation}
where \(g=\lceil \frac{j}{D} \rceil\), \(W_{K_{b, j}}\) and \(W_{V_{b, j}}\) are the original projection matrices, \(\tilde {W}_{K_{b, g}}\) and \(\tilde {W}_{V_{b, g}}\) are the newly added projection matrices, \(W_{K_{b, j}}^{apply}\) and \(W_{V_{b, j}}^{apply}\) are the actual projection matrices employed during pruning. Following the \(\mathit{L_0}\) regularization approach, we parametrize the pruning masks to hard concrete distributions. Initially, each mask is set $z=1$, we constrain the masks to zero during pruning \cite{xia2023sheared}. And the original projection matrix will be transferred to the new matrix when $z=0$. Unlike traditional $\mathit{L_0}$ regularization, we aim to eliminate any original key or value heads and just utilize $\mathit{L_0}$ masks to gradually transfer the original heads to newly added heads. All masks across blocks are constrained by a single loss function:

\begin{scriptsize}
\begin{equation}
\tilde{\mathcal{L}}_{L_0}=|(\frac{1}{N_{block} H}\sum z)-T|+\left((\frac{1}{N_{block} H}\sum z)-T\right)^2
\end{equation}
\end{scriptsize}
where \(T\) is the target size and equals zero after sparsity warm-up steps.

Before pruning, we already SFT an MHA model as teacher model. Then we use vanilla KL loss and BiLD loss \cite{li2024bild} to encourage alignment of student logits with teacher logits. 
\begin{equation}
\tilde{\mathcal{L}}_{distill} =\tilde{\mathcal{L}}_{KL}+\tilde{\mathcal{L}}_{BiLD}
\end{equation}
To sum up, the overall training loss is:
\begin{equation}
\mathcal{L} =\tilde{\mathcal{L}}_{distill}+\tilde{\mathcal{L}}_{L_0}
\end{equation}

\section{Experiments}
\subsection{Settings}
\textbf{Model configurations.} We apply our method to the LLaMA2-7B model \cite{touvron2023llama} and Sheared-LLaMA-1.3B \cite{xia2023sheared} throughout all experiments. We will prune the KV heads of MHA at different pruning rates, and compare them to the fine-tuned MHA model separately.

\textbf{Datasets.} We use the following open-source datasets for training and evaluation: BoolQ \cite{clark2019boolq}, PIQA \cite{bisk2020piqa}, HellaSwag \cite{zellers2019hellaswag}, WinoGrande \cite{sakaguchi2021winogrande}, ARC-easy \cite{clark2018think}, ARC-challenge\cite{clark2018think}, SIQA \cite{sap2019socialiqa} and OpenbookQA \cite{mihaylov2018can}. The size and instruction template of each dataset are listed in Appendix~\ref{sec:appendixB}. 

\textbf{Implementation Details.} We use 1 A100 GPU to perform model transformation, and 8 A100 GPUs for SFT the teacher model and pruning training. We randomly select 128 sequences of 2048 tokens long from the C4 training set \cite{raffel2020exploring} as calibration data in attention heads transformation. During the transformation, we convert the model parameters to float64 to reduce the calculation error. The time required for calibration and conversion is detailed in Appendix~\ref{sec:appendixC}. In pruning training, the initial learning rate is 1e-5 for the model parameters and 1e-2 for the pruning masks. The cosine scheduler is employed to reduce the learning rate to 0 by the end of training. More hyperparameter settings can be found in Appendix~\ref{sec:appendixA}. 

\begin{table*}[ht]
  \centering
  \resizebox{\textwidth}{!}{
  \renewcommand\arraystretch{1.2}
  \begin{tabular}{c|llccccccccc|c}
    \hline
    \textbf{Model}&\multicolumn{2}{c}{\textbf{Methods}} & BoolQ & PIQA & HellaSwag & WinoGrande & ARC-C & ARC-E & OpenbookQA & SIQA & \textbf{Avg.} & \makecell[c]{\textbf{Budgets}\\(tokens and epochs)}\\
    \hline
    MHA & \multicolumn{2}{c}{Teacher} & 84.83& 72.74& 68.31& 71.35& 53.18& 64.04& 59.20& 71.34& 71.37& \makecell[c]{55.8M\\3 epochs}\\
    \hline
    \hline
    \multirow{7}{*}{GQA-8}&\multicolumn{2}{c}{baseline} & 81.99& 69.10& 58.22& 71.51& 44.82& 61.05& 60.20& 68.73& 64.99& \multirow{7}{*}{\makecell[c]{111.6M\\6 epochs}}\\
    \cline{2-12}
    &\multirow{2}{*}{default grouping} &cos & 83.27& 70.95& 63.60& 71.43& \textbf{50.50}& 62.63& 58.60& 70.52& 68.38&\\
    & &dist &\textbf{83.85}&71.11&63.54&71.35&\textbf{50.50}&\textbf{63.68}&58.60&\textbf{71.03}&\textbf{68.54}&\\
    \cline{2-12}
    &\multirow{2}{*}{grouping by key} &cos &82.72& 69.75& 62.42& 70.96& 46.49& 61.58& 59.60& 68.68& 67.29&\\
    & &dist &82.11&65.34&62.61&69.38&40.80&54.21&49.80&66.22&65.99&\\
    \cline{2-12}
    &\multirow{2}{*}{grouping by value} &cos & 83.06& 70.35& \textbf{63.63}& 72.14&49.83&61.75&60.40&70.32&68.33&\\
    & &dist &83.55&\textbf{71.21}&63.54&\textbf{73.56}&46.82&61.75&\textbf{60.60}&70.57&68.53&\\
    \hline
    \hline
    \multirow{7}{*}{GQA-4}&\multicolumn{2}{c}{baseline} &79.33&64.53&50.42&70.64&38.13&57.02&54.80&66.79&59.55 &\multirow{7}{*}{\makecell[c]{223.2M\\12 epochs}}\\
    \cline{2-12}
    &\multirow{2}{*}{default grouping} &cos &81.41&68.93&59.30&\textbf{71.98}&42.14&57.54&56.00&69.04&65.24&\\
    & &dist & 81.10& \textbf{69.20}& 59.47& 70.80&43.81&59.65&57.20&68.89& 65.33&\\
    \cline{2-12}
    &\multirow{2}{*}{grouping by key} &cos &81.99&68.23&59.94&71.43&45.15&58.60&59.00&68.83&65.69&\\
    & &dist &81.50&67.95&59.54&71.35&\textbf{47.83}&57.54&59.80&67.96&65.32&\\
    \cline{2-12}
    &\multirow{2}{*}{grouping by value} &cos &81.25& 68.28& \textbf{60.06}& 70.32& 47.16& \textbf{60.18}& \textbf{60.40}& \textbf{69.14}& \textbf{65.71}&\\
    & &dist &\textbf{82.29}&67.63& 59.77& 70.40& 46.82& 60.00& 60.20& \textbf{69.14}& 65.66&\\
    \hline
  \end{tabular}
  } 
  \caption{
    Performances of Sheared-LLaMA-1.3B with various methods.
    \label{tab:results2}
  }
\end{table*}
\subsection{Ablation studies}
We test the impact of different similarity evaluation criteria (see Section \ref{subsec:standard}) and grouping strategies (see Section \ref{subsec:grouping}). All results are presented in Table~\ref{tab:results1} and Table~\ref{tab:results2}. Here, "baseline" refers to pruning directly without any transformation, "default grouping" refers to merging adjacent attention heads, "grouping by key" and "grouping by value" indicate grouping attention heads based on key or value cache similarity. "cos" and "dist" represent the transformation based on cosine similarity or Euclidean distance.

\subsection{Main results}
We report the experimental results and budgets in Table~\ref{tab:results1} and Table~\ref{tab:results2}, Avg. (Average Accuracy), indicates the average accuracy of all these sub datasets. Except for one set of experiments, all transformed models outperform the baseline. As the sparsity of KV head increases, the advantage of model transformation becomes more obvious, demonstrating the effectiveness of aligning attention heads.

\subsection{Analysis of the results}
Although our experiments utilize \(\mathit{L_0}\) regularization to accelerate the training process, model transformation can benefit any MHA to GQA conversion process. In certain cases, value-based grouping also contributed to result improvements. For experiments where grouping failed to yield performance gains, we hypothesize that cumulative errors introduced by the transformation may have a negative impact on model performance.
 
\begin{figure}[h]
  \includegraphics[trim=0.9cm 0cm 2cm 2cm, clip, width=\linewidth]{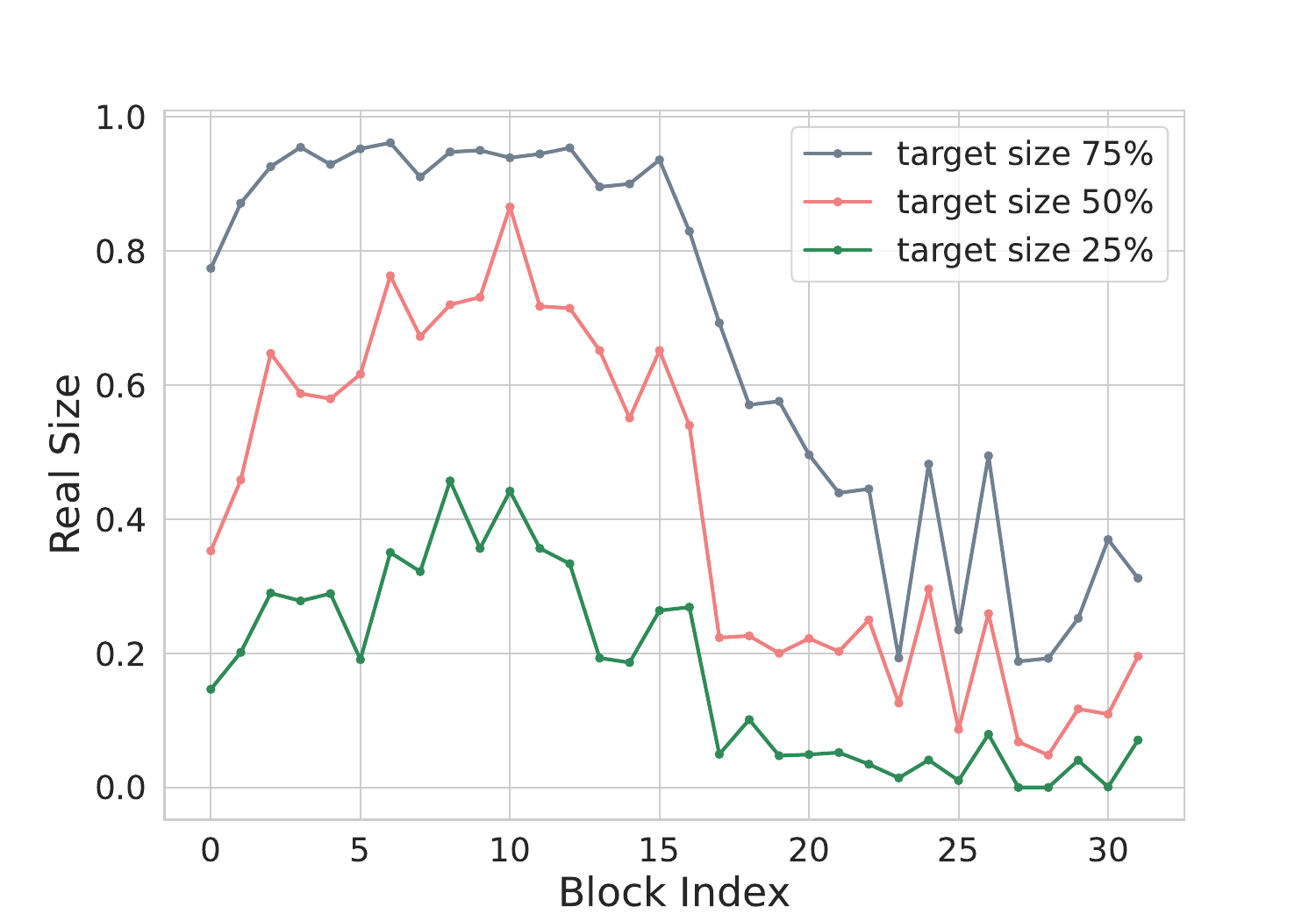}
  \caption {The shared sparsity of \(\mathit{L_0}\) masks across blocks allows different pruning speeds for different blocks in LLaMA2-7B, leading to a more stable training process.}
  \label{fig:sparsity}
\end{figure}

In addition, during the experiments, we found that the model retaining original KV heads failed to converge during the pruning process. That's why we choose not to retain any original KV heads, this setting also allows different pruning speeds for different blocks. Figure~\ref{fig:sparsity} shows the actual average size of masks in each block at different target sizes. Attention heads closer to the input layer are pruned last, as pruning these heads significantly impacts all subsequent layers. 

\section{Conclusion}
In this paper, we propose a general method for converting an MHA model into a GQA model with any compression ratio of KV heads. We find that applying orthogonal transformations to attention heads based on Procrustes analysis can enhance the similarity between KV heads without changing the model, thereby improving its performance after pruning. Furthermore, we introduce $\mathit{L_0}$ regularization during pruning training, which reduces the impact of directly eliminating parameters on the model. Our method is applicable to all KV head pruning conditions.

\section*{Limitations}
Our work has two main limitations. First, we do not delve into the grouping method, and the current approach can be further optimized. Identifying a more effective grouping strategy is one of the potential directions for future research. Moreover, our method entirely relies on the statistical mathematical features of attention heads, without considering their semantic information. In fact, compressing attention heads based on semantic information is also a promising direction \cite{tang2024razorattention}.


\begin{thebibliography}{46}
\providecommand{\natexlab}[1]{#1}
\providecommand{\url}[1]{\texttt{#1}}
\expandafter\ifx\csname urlstyle\endcsname\relax
  \providecommand{\doi}[1]{doi: #1}\else
  \providecommand{\doi}{doi: \begingroup \urlstyle{rm}\Url}\fi

\bibitem[Adnan et~al.(2024)Adnan, Arunkumar, Jain, Nair, Soloveychik, and Kamath]{adnan2024keyformer}
Muhammad Adnan, Akhil Arunkumar, Gaurav Jain, Prashant Nair, Ilya Soloveychik, and Purushotham Kamath.
\newblock Keyformer: Kv cache reduction through key tokens selection for efficient generative inference.
\newblock \emph{Proceedings of Machine Learning and Systems}, 6:\penalty0 114--127, 2024.

\bibitem[Ainslie et~al.(2023)Ainslie, Lee-Thorp, de~Jong, Zemlyanskiy, Lebr{\'o}n, and Sanghai]{ainslie2023gqa}
Joshua Ainslie, James Lee-Thorp, Michiel de~Jong, Yury Zemlyanskiy, Federico Lebr{\'o}n, and Sumit Sanghai.
\newblock Gqa: Training generalized multi-query transformer models from multi-head checkpoints.
\newblock \emph{arXiv preprint arXiv:2305.13245}, 2023.

\bibitem[Ashkboos et~al.(2024)Ashkboos, Croci, Nascimento, Hoefler, and Hensman]{ashkboos2024slicegpt}
Saleh Ashkboos, Maximilian~L Croci, Marcelo Gennari~do Nascimento, Torsten Hoefler, and James Hensman.
\newblock Slicegpt: Compress large language models by deleting rows and columns.
\newblock \emph{arXiv preprint arXiv:2401.15024}, 2024.

\bibitem[Bisk et~al.(2020)Bisk, Zellers, Gao, Choi, et~al.]{bisk2020piqa}
Yonatan Bisk, Rowan Zellers, Jianfeng Gao, Yejin Choi, et~al.
\newblock Piqa: Reasoning about physical commonsense in natural language.
\newblock In \emph{Proceedings of the AAAI conference on artificial intelligence}, volume~34, pages 7432--7439, 2020.

\bibitem[Brown et~al.(2020)Brown, Mann, Ryder, Subbiah, Kaplan, Dhariwal, Neelakantan, Shyam, Sastry, Askell, Agarwal, Herbert-Voss, Krueger, Henighan, Child, Ramesh, Ziegler, Wu, Winter, Hesse, Chen, Sigler, Litwin, Gray, Chess, Clark, Berner, McCandlish, Radford, Sutskever, and Amodei]{NEURIPS2020_1457c0d6}
Tom Brown, Benjamin Mann, Nick Ryder, Melanie Subbiah, Jared~D Kaplan, Prafulla Dhariwal, Arvind Neelakantan, Pranav Shyam, Girish Sastry, Amanda Askell, Sandhini Agarwal, Ariel Herbert-Voss, Gretchen Krueger, Tom Henighan, Rewon Child, Aditya Ramesh, Daniel Ziegler, Jeffrey Wu, Clemens Winter, Chris Hesse, Mark Chen, Eric Sigler, Mateusz Litwin, Scott Gray, Benjamin Chess, Jack Clark, Christopher Berner, Sam McCandlish, Alec Radford, Ilya Sutskever, and Dario Amodei.
\newblock Language models are few-shot learners.
\newblock In H.~Larochelle, M.~Ranzato, R.~Hadsell, M.F. Balcan, and H.~Lin, editors, \emph{Advances in Neural Information Processing Systems}, volume~33, pages 1877--1901. Curran Associates, Inc., 2020.

\bibitem[Cai et~al.(2013)Cai, Fan, and Jiang]{cai2013distributions}
Tony Cai, Jianqing Fan, and Tiefeng Jiang.
\newblock Distributions of angles in random packing on spheres.
\newblock \emph{The Journal of Machine Learning Research}, 14\penalty0 (1):\penalty0 1837--1864, 2013.

\bibitem[Chen et~al.(2024{\natexlab{a}})Chen, Zhang, Shang, Zhang, Liu, Wang, and Sun]{chen2024dha}
Yilong Chen, Linhao Zhang, Junyuan Shang, Zhenyu Zhang, Tingwen Liu, Shuohuan Wang, and Yu~Sun.
\newblock Dha: Learning decoupled-head attention from transformer checkpoints via adaptive heads fusion.
\newblock \emph{arXiv preprint arXiv:2406.06567}, 2024{\natexlab{a}}.

\bibitem[Chen et~al.(2024)Chen, Zhang, Gao, Mullins, Constantinides, and Zhao]{chen2024optimised}
Yuang Chen, Cheng Zhang, Xitong Gao, Robert~D Mullins, George~A Constantinides, and Yiren Zhao.
\newblock Optimised grouped-query attention mechanism for transformers.
\newblock \emph{arXiv preprint arXiv:2406.14963}, 2024.

\bibitem[Clark et~al.(2019)Clark, Lee, Chang, Kwiatkowski, Collins, and Toutanova]{clark2019boolq}
Christopher Clark, Kenton Lee, Ming-Wei Chang, Tom Kwiatkowski, Michael Collins, and Kristina Toutanova.
\newblock Boolq: Exploring the surprising difficulty of natural yes/no questions.
\newblock \emph{arXiv preprint arXiv:1905.10044}, 2019.

\bibitem[Clark et~al.(2018)Clark, Cowhey, Etzioni, Khot, Sabharwal, Schoenick, and Tafjord]{clark2018think}
Peter Clark, Isaac Cowhey, Oren Etzioni, Tushar Khot, Ashish Sabharwal, Carissa Schoenick, and Oyvind Tafjord.
\newblock Think you have solved question answering? try arc, the ai2 reasoning challenge.
\newblock \emph{arXiv preprint arXiv:1803.05457}, 2018.

\bibitem[Dubey et~al.(2024)Dubey, Jauhri, Pandey, Kadian, Al-Dahle, Letman, Mathur, Schelten, Yang, Fan, et~al.]{dubey2024llama}
Abhimanyu Dubey, Abhinav Jauhri, Abhinav Pandey, Abhishek Kadian, Ahmad Al-Dahle, Aiesha Letman, Akhil Mathur, Alan Schelten, Amy Yang, Angela Fan, et~al.
\newblock The llama 3 herd of models.
\newblock \emph{arXiv preprint arXiv:2407.21783}, 2024.

\bibitem[Frantar and Alistarh(2023)]{frantar2023sparsegpt}
Elias Frantar and Dan Alistarh.
\newblock Sparsegpt: Massive language models can be accurately pruned in one-shot.
\newblock In \emph{International Conference on Machine Learning}, pages 10323--10337. PMLR, 2023.

\bibitem[Gou et~al.(2021)Gou, Yu, Maybank, and Tao]{gou2021knowledge}
Jianping Gou, Baosheng Yu, Stephen~J Maybank, and Dacheng Tao.
\newblock Knowledge distillation: A survey.
\newblock \emph{International Journal of Computer Vision}, 129\penalty0 (6):\penalty0 1789--1819, 2021.

\bibitem[Hendrycks et~al.(2020)Hendrycks, Burns, Basart, Zou, Mazeika, Song, and Steinhardt]{hendrycks2020measuring}
Dan Hendrycks, Collin Burns, Steven Basart, Andy Zou, Mantas Mazeika, Dawn Song, and Jacob Steinhardt.
\newblock Measuring massive multitask language understanding.
\newblock \emph{arXiv preprint arXiv:2009.03300}, 2020.

\bibitem[Hooper et~al.(2024)Hooper, Kim, Mohammadzadeh, Mahoney, Shao, Keutzer, and Gholami]{hooper2024kvquant}
Coleman Richard~Charles Hooper, Sehoon Kim, Hiva Mohammadzadeh, Michael~W. Mahoney, Sophia Shao, Kurt Keutzer, and Amir Gholami.
\newblock {KVQ}uant: Towards 10 million context length {LLM} inference with {KV} cache quantization.
\newblock In \emph{The Thirty-eighth Annual Conference on Neural Information Processing Systems}, 2024.
\newblock URL \url{https://openreview.net/forum?id=0LXotew9Du}.

\bibitem[Hu et~al.(2021)Hu, Shen, Wallis, Allen-Zhu, Li, Wang, Wang, and Chen]{hu2021lora}
Edward~J Hu, Yelong Shen, Phillip Wallis, Zeyuan Allen-Zhu, Yuanzhi Li, Shean Wang, Lu~Wang, and Weizhu Chen.
\newblock Lora: Low-rank adaptation of large language models.
\newblock \emph{arXiv preprint arXiv:2106.09685}, 2021.

\bibitem[Jiang et~al.(2023)Jiang, Sablayrolles, Mensch, Bamford, Chaplot, Casas, Bressand, Lengyel, Lample, Saulnier, et~al.]{jiang2023mistral}
Albert~Q Jiang, Alexandre Sablayrolles, Arthur Mensch, Chris Bamford, Devendra~Singh Chaplot, Diego de~las Casas, Florian Bressand, Gianna Lengyel, Guillaume Lample, Lucile Saulnier, et~al.
\newblock Mistral 7b.
\newblock \emph{arXiv preprint arXiv:2310.06825}, 2023.

\bibitem[Li et~al.(2024)Li, Zhou, and Song]{li2024bild}
Minchong Li, Feng Zhou, and Xiaohui Song.
\newblock Bild: Bi-directional logits difference loss for large language model distillation.
\newblock \emph{arXiv preprint arXiv:2406.13555}, 2024.

\bibitem[Liu et~al.(2024{\natexlab{a}})Liu, Feng, Wang, Wang, Liu, Zhao, Dengr, Ruan, Dai, Guo, et~al.]{liu2024deepseek-v2}
Aixin Liu, Bei Feng, Bin Wang, Bingxuan Wang, Bo~Liu, Chenggang Zhao, Chengqi Dengr, Chong Ruan, Damai Dai, Daya Guo, et~al.
\newblock Deepseek-v2: A strong, economical, and efficient mixture-of-experts language model.
\newblock \emph{arXiv preprint arXiv:2405.04434}, 2024{\natexlab{a}}.

\bibitem[Liu et~al.(2024{\natexlab{b}})Liu, Feng, Xue, Wang, Wu, Lu, Zhao, Deng, Zhang, Ruan, et~al.]{liu2024deepseek-v3}
Aixin Liu, Bei Feng, Bing Xue, Bingxuan Wang, Bochao Wu, Chengda Lu, Chenggang Zhao, Chengqi Deng, Chenyu Zhang, Chong Ruan, et~al.
\newblock Deepseek-v3 technical report.
\newblock \emph{arXiv preprint arXiv:2412.19437}, 2024{\natexlab{b}}.

\bibitem[Liu et~al.(2024{\natexlab{c}})Liu, Zhang, Li, Yan, Gao, Chen, Yuan, Huang, Sun, Gao, et~al.]{liu2024sora}
Yixin Liu, Kai Zhang, Yuan Li, Zhiling Yan, Chujie Gao, Ruoxi Chen, Zhengqing Yuan, Yue Huang, Hanchi Sun, Jianfeng Gao, et~al.
\newblock Sora: A review on background, technology, limitations, and opportunities of large vision models.
\newblock \emph{arXiv preprint arXiv:2402.17177}, 2024{\natexlab{c}}.

\bibitem[Liu et~al.(2023{\natexlab{a}})Liu, Desai, Liao, Wang, Xie, Xu, Kyrillidis, and Shrivastava]{liu2023scissorhands}
Zichang Liu, Aditya Desai, Fangshuo Liao, Weitao Wang, Victor Xie, Zhaozhuo Xu, Anastasios Kyrillidis, and Anshumali Shrivastava.
\newblock Scissorhands: Exploiting the persistence of importance hypothesis for llm kv cache compression at test time.
\newblock \emph{Advances in Neural Information Processing Systems}, 36:\penalty0 52342--52364, 2023{\natexlab{a}}.

\bibitem[Liu et~al.(2023{\natexlab{b}})Liu, Wang, Dao, Zhou, Yuan, Song, Shrivastava, Zhang, Tian, Re, et~al.]{liu2023deja}
Zichang Liu, Jue Wang, Tri Dao, Tianyi Zhou, Binhang Yuan, Zhao Song, Anshumali Shrivastava, Ce~Zhang, Yuandong Tian, Christopher Re, et~al.
\newblock Deja vu: Contextual sparsity for efficient llms at inference time.
\newblock In \emph{International Conference on Machine Learning}, pages 22137--22176. PMLR, 2023{\natexlab{b}}.

\bibitem[Longpre et~al.(2023)Longpre, Hou, Vu, Webson, Chung, Tay, Zhou, Le, Zoph, Wei, et~al.]{longpre2023flan}
Shayne Longpre, Le~Hou, Tu~Vu, Albert Webson, Hyung~Won Chung, Yi~Tay, Denny Zhou, Quoc~V Le, Barret Zoph, Jason Wei, et~al.
\newblock The flan collection: Designing data and methods for effective instruction tuning.
\newblock In \emph{International Conference on Machine Learning}, pages 22631--22648. PMLR, 2023.

\bibitem[Louizos et~al.(2017)Louizos, Welling, and Kingma]{louizos2017learning}
Christos Louizos, Max Welling, and Diederik~P Kingma.
\newblock Learning sparse neural networks through $ l\_0 $ regularization.
\newblock \emph{arXiv preprint arXiv:1712.01312}, 2017.

\bibitem[Mihaylov et~al.(2018)Mihaylov, Clark, Khot, and Sabharwal]{mihaylov2018can}
Todor Mihaylov, Peter Clark, Tushar Khot, and Ashish Sabharwal.
\newblock Can a suit of armor conduct electricity? a new dataset for open book question answering.
\newblock \emph{arXiv preprint arXiv:1809.02789}, 2018.

\bibitem[Noach and Goldberg(2020)]{noach2020compressing}
Matan~Ben Noach and Yoav Goldberg.
\newblock Compressing pre-trained language models by matrix decomposition.
\newblock In \emph{Proceedings of the 1st Conference of the Asia-Pacific Chapter of the Association for Computational Linguistics and the 10th International Joint Conference on Natural Language Processing}, pages 884--889, 2020.

\bibitem[Ouyang et~al.(2022)Ouyang, Wu, Jiang, Almeida, Wainwright, Mishkin, Zhang, Agarwal, Slama, Ray, et~al.]{ouyang2022training}
Long Ouyang, Jeffrey Wu, Xu~Jiang, Diogo Almeida, Carroll Wainwright, Pamela Mishkin, Chong Zhang, Sandhini Agarwal, Katarina Slama, Alex Ray, et~al.
\newblock Training language models to follow instructions with human feedback.
\newblock \emph{Advances in neural information processing systems}, 35:\penalty0 27730--27744, 2022.

\bibitem[Radford et~al.(2018)Radford, Narasimhan, Salimans, and Sutskever]{radford2018improving}
Alec Radford, Karthik Narasimhan, Tim Salimans, and Ilya Sutskever.
\newblock Improving language understanding by generative pre-training.
\newblock 2018.

\bibitem[Raffel et~al.(2020)Raffel, Shazeer, Roberts, Lee, Narang, Matena, Zhou, Li, and Liu]{raffel2020exploring}
Colin Raffel, Noam Shazeer, Adam Roberts, Katherine Lee, Sharan Narang, Michael Matena, Yanqi Zhou, Wei Li, and Peter~J Liu.
\newblock Exploring the limits of transfer learning with a unified text-to-text transformer.
\newblock \emph{Journal of machine learning research}, 21\penalty0 (140):\penalty0 1--67, 2020.

\bibitem[Sakaguchi et~al.(2021)Sakaguchi, Bras, Bhagavatula, and Choi]{sakaguchi2021winogrande}
Keisuke Sakaguchi, Ronan~Le Bras, Chandra Bhagavatula, and Yejin Choi.
\newblock Winogrande: An adversarial winograd schema challenge at scale.
\newblock \emph{Communications of the ACM}, 64\penalty0 (9):\penalty0 99--106, 2021.


\bibitem[Sap et~al.(2019)Sap, Rashkin, Chen, LeBras, and Choi]{sap2019socialiqa}
Maarten Sap, Hannah Rashkin, Derek Chen, Ronan LeBras, and Yejin Choi.
\newblock Socialiqa: Commonsense reasoning about social interactions.
\newblock \emph{arXiv preprint arXiv:1904.09728}, 2019.

\bibitem[Sch{\"o}nemann(1966)]{schonemann1966generalized}
Peter~H Sch{\"o}nemann.
\newblock A generalized solution of the orthogonal procrustes problem.
\newblock \emph{Psychometrika}, 31\penalty0 (1):\penalty0 1--10, 1966.

\bibitem[Shazeer(2019)]{shazeer2019fast}
Noam Shazeer.
\newblock Fast transformer decoding: One write-head is all you need.
\newblock \emph{arXiv preprint arXiv:1911.02150}, 2019.

\bibitem[Socher et~al.(2013)Socher, Perelygin, Wu, Chuang, Manning, Ng, and Potts]{socher2013recursive}
Richard Socher, Alex Perelygin, Jean Wu, Jason Chuang, Christopher~D Manning, Andrew~Y Ng, and Christopher Potts.
\newblock Recursive deep models for semantic compositionality over a sentiment treebank.
\newblock In \emph{Proceedings of the 2013 conference on empirical methods in natural language processing}, pages 1631--1642, 2013.

\bibitem[Su et~al.(2024)Su, Ahmed, Lu, Pan, Bo, and Liu]{su2024roformer}
Jianlin Su, Murtadha Ahmed, Yu~Lu, Shengfeng Pan, Wen Bo, and Yunfeng Liu.
\newblock Roformer: Enhanced transformer with rotary position embedding.
\newblock \emph{Neurocomputing}, 568:\penalty0 127063, 2024.

\bibitem[Sun et~al.(2020)Sun, Yu, Yu, and Cardie]{sun2020investigating}
Kai Sun, Dian Yu, Dong Yu, and Claire Cardie.
\newblock Investigating prior knowledge for challenging chinese machine reading comprehension.
\newblock \emph{Transactions of the Association for Computational Linguistics}, 8:\penalty0 141--155, 2020.

\bibitem[Sun et~al.(2023)Sun, Liu, Bair, and Kolter]{sun2023simple}
Mingjie Sun, Zhuang Liu, Anna Bair, and J~Zico Kolter.
\newblock A simple and effective pruning approach for large language models.
\newblock \emph{arXiv preprint arXiv:2306.11695}, 2023.

\bibitem[Tang et~al.(2024)Tang, Lin, Lin, Han, Hong, Yao, and Wang]{tang2024razorattention}
Hanlin Tang, Yang Lin, Jing Lin, Qingsen Han, Shikuan Hong, Yiwu Yao, and Gongyi Wang.
\newblock Razorattention: Efficient kv cache compression through retrieval heads.
\newblock \emph{arXiv preprint arXiv:2407.15891}, 2024.

\bibitem[Touvron et~al.(2023)Touvron, Martin, Stone, Albert, Almahairi, Babaei, Bashlykov, Batra, Bhargava, Bhosale, et~al.]{touvron2023llama}
Hugo Touvron, Louis Martin, Kevin Stone, Peter Albert, Amjad Almahairi, Yasmine Babaei, Nikolay Bashlykov, Soumya Batra, Prajjwal Bhargava, Shruti Bhosale, et~al.
\newblock Llama 2: Open foundation and fine-tuned chat models.
\newblock \emph{arXiv preprint arXiv:2307.09288}, 2023.

\bibitem[Vaswani et~al.(2017)Vaswani, Shazeer, Parmar, Uszkoreit, Jones, Gomez, Kaiser, and Polosukhin]{NIPS2017_3f5ee243}
Ashish Vaswani, Noam Shazeer, Niki Parmar, Jakob Uszkoreit, Llion Jones, Aidan~N Gomez, \L~ukasz Kaiser, and Illia Polosukhin.
\newblock Attention is all you need.
\newblock In I.~Guyon, U.~Von Luxburg, S.~Bengio, H.~Wallach, R.~Fergus, S.~Vishwanathan, and R.~Garnett, editors, \emph{Advances in Neural Information Processing Systems}, volume~30. Curran Associates, Inc., 2017.
\newblock URL \url{https://proceedings.neurips.cc/paper_files/paper/2017/file/3f5ee243547dee91fbd053c1c4a845aa-Paper.pdf}.

\bibitem[Wang et~al.(2019)Wang, Wohlwend, and Lei]{wang2019structured}
Ziheng Wang, Jeremy Wohlwend, and Tao Lei.
\newblock Structured pruning of large language models.
\newblock \emph{arXiv preprint arXiv:1910.04732}, 2019.

\bibitem[{Wikipedia contributors}(2022)]{enwiki:1126373270}
{Wikipedia contributors}.
\newblock Generalized procrustes analysis --- {Wikipedia}{,} the free encyclopedia, 2022.
\newblock URL \url{https://en.wikipedia.org/w/index.php?title=Generalized_Procrustes_analysis&oldid=1126373270}.
\newblock [Online; accessed 24-October-2024].

\bibitem[Xia et~al.(2022)Xia, Zhong, and Chen]{xia2022structured}
Mengzhou Xia, Zexuan Zhong, and Danqi Chen.
\newblock Structured pruning learns compact and accurate models.
\newblock \emph{arXiv preprint arXiv:2204.00408}, 2022.

\bibitem[Xia et~al.(2023)Xia, Gao, Zeng, and Chen]{xia2023sheared}
Mengzhou Xia, Tianyu Gao, Zhiyuan Zeng, and Danqi Chen.
\newblock Sheared llama: Accelerating language model pre-training via structured pruning.
\newblock \emph{arXiv preprint arXiv:2310.06694}, 2023.


\bibitem[Yang et~al.(2024{\natexlab{a}})Yang, Yang, Hui, Zheng, Yu, Zhou, Li, Li, Liu, Huang, et~al.]{yang2024qwen2}
An~Yang, Baosong Yang, Binyuan Hui, Bo~Zheng, Bowen Yu, Chang Zhou, Chengpeng Li, Chengyuan Li, Dayiheng Liu, Fei Huang, et~al.
\newblock Qwen2 technical report.
\newblock \emph{arXiv preprint arXiv:2407.10671}, 2024{\natexlab{a}}.

\bibitem[Yang et~al.(2024{\natexlab{b}})Yang, Kim, Bae, Kwon, Park, Yang, Kwon, and Lee]{yang2024no}
June~Yong Yang, Byeongwook Kim, Jeongin Bae, Beomseok Kwon, Gunho Park, Eunho Yang, Se~Jung Kwon, and Dongsoo Lee.
\newblock No token left behind: Reliable kv cache compression via importance-aware mixed precision quantization.
\newblock \emph{arXiv preprint arXiv:2402.18096}, 2024{\natexlab{b}}.

\bibitem[Yu and Wu(2023)]{yu2023compressing}
Hao Yu and Jianxin Wu.
\newblock Compressing transformers: features are low-rank, but weights are not!
\newblock In \emph{Proceedings of the AAAI Conference on Artificial Intelligence}, volume~37, pages 11007--11015, 2023.

\bibitem[Yu et~al.(2024)Yu, Yang, Li, Li, and Wu]{yu2024effectively}
Hao Yu, Zelan Yang, Shen Li, Yong Li, and Jianxin Wu.
\newblock Effectively compress kv heads for llm.
\newblock \emph{arXiv preprint arXiv:2406.07056}, 2024.

\bibitem[Yue et~al.(2024)Yue, Yuan, Duanmu, Zhou, Wu, and Nie]{yue2024wkvquant}
Yuxuan Yue, Zhihang Yuan, Haojie Duanmu, Sifan Zhou, Jianlong Wu, and Liqiang Nie.
\newblock Wkvquant: Quantizing weight and key/value cache for large language models gains more.
\newblock \emph{arXiv preprint arXiv:2402.12065}, 2024.

\bibitem[Zellers et~al.(2019)Zellers, Holtzman, Bisk, Farhadi, and Choi]{zellers2019hellaswag}
Rowan Zellers, Ari Holtzman, Yonatan Bisk, Ali Farhadi, and Yejin Choi.
\newblock Hellaswag: Can a machine really finish your sentence?
\newblock \emph{arXiv preprint arXiv:1905.07830}, 2019.

\bibitem[Zhang et~al.(2024)Zhang, Zeng, Wang, and Lu]{zhang2024tinyllama}
Peiyuan Zhang, Guangtao Zeng, Tianduo Wang, and Wei Lu.
\newblock Tinyllama: An open-source small language model.
\newblock \emph{arXiv preprint arXiv:2401.02385}, 2024.

\end{thebibliography}

\onecolumn
\appendix

\section{Hyperparameter settings}
To reduce memory usage, we employ DeepSpeed during both SFT and pruning training, we set \(k\)=16 for BiLD loss \cite{li2024bild}. During the pruning training process, the sparsity warm-up steps account for 30\% of the total steps, during which the target size of the $\mathit{L_0}$ masks decreases linearly to zero. The maximum pruning steps comprise 80\% of the total steps, after which the mask training ceases, only the model parameters are adjusted. Some more hyperparameter settings for SFT teacher model and pruning training are shown in Table~\ref{tab:hyperparameters}. 
\label{sec:appendixA}

\begin{table}[h]
\centering
\small
\renewcommand\arraystretch{1.5}
\begin{tabular}{lcc}
\toprule
&SFT teacher & pruning training\\
\hline
batch size & 128 & 64\\
\hline
micro batch size & 4 & 1\\
\hline
lr warmup steps& 16 & 32 \\
\hline
initial lr of masks& \textbackslash &1e-2\\
\hline
initial lr of model& \multicolumn{2}{c}{1e-5}  \\
\bottomrule
\end{tabular}
\caption{
    Some hyperparameters setting for experiments.
    \label{tab:hyperparameters}
  }
\end{table}
\section{Details of datasets}
\label{sec:appendixB}
The sizes of sub datasets are shown in Table ~\ref{tab:size}. 
\begin{table}[h]
    \centering
    \begin{tabular}{c|cc}
      \textbf{\makecell[c]{datasets}} & train & test \\ 
      \hline
      BoolQ & 9427 & 3270 \\
      PIQA & 16113 & 1838  \\
      HellaSwag & 39905 & 10042 \\
      WinoGrande & 40398 & 1267 \\
      ARC-C & 1119 & 299 \\
      ARC-E & 2251 & 570 \\
      OpenbookQA & 4957 & 500 \\
      SIQA & 33410 & 1954 \\
      \hline
      total & 147580 & 19740\\
      
    \end{tabular}   
    \caption{Sizes of different datasets
        \label{tab:size}
    }
\end{table}

The template of each dataset can be seen in Table ~\ref{tab:templates}.

\clearpage
\begin{longtable}[t]{|c|p{12.5cm}|}
\hline
Dataset & Template \\
\hline
\makecell[c]{Arc-C \\ Arc-E \\OpenbookQA}&\makecell[l]{Which color shirt will reflect the most light on a hot, sunny day? \\ Choices: ['black', 'blue', 'red', 'white'] \\Answer:}\\
\hline
\makecell[c]{HellaSwag} &{Please choose the most appropriate text to complete the passage below: \newline Passage: A male athlete puts powder on his hands. he \newline Choices: ['bends and inspects his hands for damage.', 'shakes them shakingly before putting them in his mouth.', 'mounts a high beam in the gym.', 'then jumps up and does a high jump.'] \newline Answer:}\\ 
\hline
\makecell[c]{BoolQ} & The Coroner -- The BBC announced on 2 March 2017 that there would be no further series.\newline
Question: will there be a second series of the coroner?\newline
Choices: ['true', 'false'] \newline
Answer: \\
\hline
\makecell[c]{Winogrande} & Choose the most sensible text to replace the '\_' in the following sentence: Natalie was less religous than Patricia, therefore \_ attended church services more often on Sundays. \newline
Choices: ['Natalie', 'Patricia'] \newline
Answer: \\
\hline
\makecell[c]{PIQA} & Goal: how do you flood a room? \newline
Choose the most sensible solution to achieve the goal. Choices: ['fill it with objects.', 'fill it with water.'] \newline
Answer: \\
\hline
\makecell[c]{SIQA\\} & Sasha took him to vegas for a vacation.\newline
Question: How would Sasha feel afterwards?? \newline
Choices: ['sad', 'depressed', 'fulfilled'] \newline
Answer: \\
\hline
\caption{The template of each dataset 
\label{tab:templates}
}
\end{longtable}

\section{Time cost of model calibration and transformation}
\label{sec:appendixC}
\begin{table}[h]
    \centering
    \begin{tabular}{cccc}   
       \multicolumn{2}{c}{\textbf{model}} & LLaMA-2-7B & Sheared-llama-1.3B\\
       \hline
       \multicolumn{2}{c}{calibration} & 5min & 3min \\
       \hline
        \multirow{2}{*}{\makecell[c]{GQA16 \\transformation}}&w/ grouping&22min &\textbackslash\\
        \cline{2-4}
       &w/o grouping&4min&\textbackslash\\
       \hline
        \multirow{2}{*}{\makecell[c]{GQA8 \\transformation}}&w/ grouping&37min &20min\\
        \cline{2-4}
       &w/o grouping&17min&2min\\
       \hline
        \multirow{2}{*}{\makecell[c]{GQA4 \\transformation}}&w/ grouping&1h2min &35min\\
        \cline{2-4}
       &w/o grouping&29min&14min\\
    \end{tabular}   
    \caption{Time cost of model calibration and transformation. It requires at most one hour to complete this procedure. All calculations are performed by 1 A100 GPU.
        \label{tab:size}
    }
\end{table}

\end{document}